\documentclass[journal]{IEEEtran}
\usepackage{amsmath,amsfonts}
\usepackage{algorithmic}
\usepackage{algorithm}
\usepackage{array}
\usepackage[caption=false,font=normalsize,labelfont=sf,textfont=sf]{subfig}
\usepackage{textcomp}
\usepackage{stfloats}
\usepackage{url}
\usepackage{verbatim}
\usepackage{graphicx}
\usepackage{cite}
\usepackage{threeparttable}
\usepackage{multirow}
\usepackage{times}
\usepackage[table]{xcolor}
\usepackage{bm}
\usepackage{soul,color}
\usepackage[hidelinks]{hyperref}
\usepackage[utf8]{inputenc}
\usepackage{amsthm}
\usepackage{booktabs}
\usepackage[switch]{lineno}

\newcommand{\ie}{\textit{i}.\textit{e}., }
\newcommand{\eg}{\textit{e}.\textit{g}., }

\begin{document}

\title{Adaptively Clustering Neighbor Elements for Image-Text Generation}

\author{Zihua~Wang,
        Xu~Yang,
        Haiyang~Xu,
        Hanwang~Zhang,
        Ming~Yan,
        Fei~Huang,
        and~Yu~Zhang*% <-this % stops a space
		\IEEEcompsocitemizethanks{
\IEEEcompsocthanksitem Zihua Wang, Xu Yang and Yu Zhang are with the School of Computer Science and Engineering, and the Key Lab of Computer Network and Information Integration (Ministry of Education), Southeast University, Nanjing 211189, China.
\IEEEcompsocthanksitem Hanwang Zhang is with the School of Computer Science and Engineering, Nanyang Technological University, Singapore.
\IEEEcompsocthanksitem Zihua Wang, Haiyang Xu, Ming Yan, and Fei Huang are with the DAMO
Academy, Alibaba Group.\protect
\IEEEcompsocthanksitem        *Corresponding author. E-mail: zhang\_yu@seu.edu.cn}
\thanks{This work has been accepted by IEEE Transactions on Multimedia.
}}%

%\IEEEpubid{~\copyright~2021 IEEE}

\maketitle

\begin{abstract}
We propose a novel Transformer-based image-to-text generation model termed as \textbf{ACF} that adaptively clusters vision patches into object regions and language words into phrases to implicitly learn object-phrase alignments for better visual-text coherence. To achieve this, we design a novel self-attention layer that applies self-attention over the elements in a local cluster window instead of the whole sequence. The window size is softly decided by a clustering matrix that is calculated by the current input data and thus this process is adaptive. By stacking these revised self-attention layers to construct ACF, the small clusters in the lower layers can be grouped into a bigger cluster, \eg vision/language. ACF clusters small objects/phrases into bigger ones. In this gradual clustering process, a parsing tree is generated which embeds the hierarchical knowledge of the input sequence. As a result, by using ACF to build the vision encoder and language decoder, the hierarchical object-phrase alignments are embedded and then transferred from vision to language domains in two popular image-to-text tasks: Image captioning and Visual Question Answering. The experiment results demonstrate the effectiveness of ACF, which outperforms most SOTA captioning and VQA models and achieves comparable scores compared with some large-scale pre-trained models. Our code is available \href{https://github.com/ZihuaEvan/ACFModel/}{[here]}.
\end{abstract}


\begin{figure}[t]
\centering
\includegraphics[width=1\linewidth,clip]{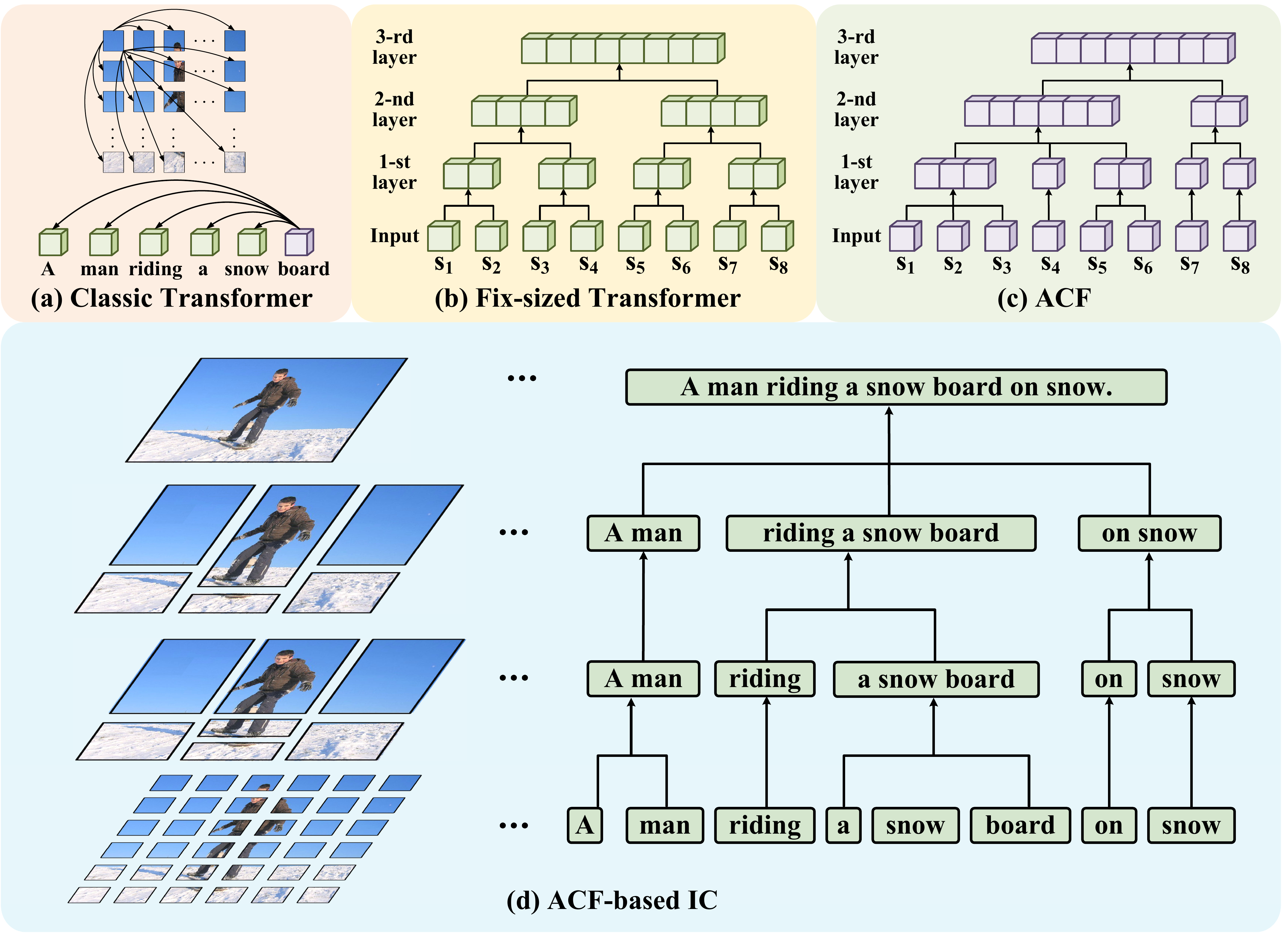}
   \caption{(a) Classic Transformer. (b) Transformer with fixed-size windows (size = $2$); (c) ACF which adjusts the window size according to the input. (d) ACF-based IC. The left/right part shows how the vision/language ACFs cluster image grids/language words for transferring structural commonalities.
   }
\label{fig:fig1}
\vspace{-0.2in}

\end{figure}

\section{Introduction}
Due to the inherent disparities between visual and textual data, researchers commonly tend to employ a unified architecture to replace the CNN+RNN/LSTM paradigm in visual-text tasks.
Transformer-based architectures have gained significant popularity in this field, finding extensive application in diverse visual-text tasks such as image captioning (IC) and visual question answering (VQA). 
The key advantage of these architectures lies in their unified design, where both the vision encoder and language decoder benefit from a shared framework. The encoder leverages Self-Attention (Self-ATT) mechanisms~\cite{vaswani2017attention} to capture visual context knowledge across all visual tokens, while the decoder intelligently selects relevant visual information based on the context of the partially generated caption or the question~\cite{herdade2019image,yu2019deep}.

However, most of the Transformers conduct the attention mechanism with fully connected graphs. Take IC for example, as shown in Figure~\ref{fig:fig1} (a), the traditional Transformer-based captioning models cannot achieve such region-phrase alignments since the Self-ATT is applied over all the input tokens.
As a result, each vision token output from the encoder embeds the context knowledge of the whole image instead of the neighbor regions. Similarly, when the decoder generates the next word, the context of all the partially generated words is used to select the suitable vision knowledge.
While intuitively, when we humans describe an image, we usually first construct suitable phrases to describe the recognized important image regions and then compose these phrases into an integral sentence.

In order to mitigate the issue of excessive connections in fully connected graphs, local-global Transformers~\cite{al2019character,liu2021swin} are proposed to implement the Self-ATT over a cluster of neighbor elements in a fixed-size small window to learn local contexts. Moreover, when stacking these cluster-constrained Self-ATT layers, the small cluster in the lower layer will be gradually merged into bigger ones for learning more global knowledge. As the result, the local-global knowledge can be learnt by these Transformer variants. For example, as shown in Figure~\ref{fig:fig1}(b), the 1-st layer clusters 2 neighboring elements like $\{\bm{s}_1,\bm{s}_2\}$ to carry Self-ATT for local contexts and the 2-nd layer merges $\{\bm{s}_1,\bm{s}_2\}$ and $\{\bm{s}_3,\bm{s}_4\}$ into a bigger one to learn more global context.

However, these global-local Transformers cannot be directly used to build captioning and VQA models to learn region-phrase alignments due to two reasons. Firstly, most previous local-global Transformers only use \textbf{fixed-size windows} to group tokens, while vision and language data have \textbf{varying graininess}, \eg objects/phrases: different numbers of grids/words at different positions, and such varying graininess cannot be learnt by these fixed-size window-based self-attentions. Secondly, to encourage an encoder-decoder to learn region-phrase alignments, the \textbf{similar inductive bias} should be applied to design both the encoder and decoder. However, most local-global Transformers~\cite{chen2021glit,liang2023local,HanFDSC0022} are exclusively designed to deal with images that exploit lots of visual inductive bias like translation invariance, which cannot be used as the language decoder.

To solve these two limitations, we propose a novel global-local Transformer which applies a general visual-linguistic inductive bias to capture varying graininess of both vision and language data. Specifically, we enable the Self-ATT layer to \textbf{Ada}ptively \textbf{Clust}er the neighbor elements for implementing self-attention and term this novel model as \textbf{Ada-ClustFormer} (\textbf{ACF}). To achieve this, we insert a probabilistic clustering matrix $\bm{C}$ into the Self-ATT layer, where the probability $\bm{C}_{ij}$ softly determines whether the sub-sequence $\{ \bm{s}_i,...,\bm{s}_j \}$ should be clustered or not. To calculate $\bm{C}_{ij}$, we consider whether the next element $\bm{s}_j$ is similar to the mean-pooling of $\{ \bm{s}_i,...,\bm{s}_{j-1} \}$. Thus we can adjust the cluster size based on each specific data sample. As shown in Figure~\ref{fig:fig1}(c), in each layer, the window size is not fixed but can be adjusted to each specific input sequence, \eg in the 1-st layer, $\{\bm{s}_1,\bm{s}_2,\bm{s}_3\}$, $\{\bm{s}_4$\}, $\{\bm{s}_5,\bm{s}_6$\}, $\{\bm{s}_7$\}, $\{\bm{s}_8\}$ are respectively clustered. 
Then ACF can be constructed by stacking these revised Self-ATT layers, while simply stacking cannot guarantee the small clusters in the lower layers to be merged into bigger ones in the higher layers. To remedy this problem, we enforce $\bm{C}^{l-1} \leq \bm{C}^{l}$, where $l$ denotes the $l$-th layer, by using a convex combination technique. Then as Figure~\ref{fig:fig1}(c) shows, the higher layers merge small clusters into bigger ones for learning global contexts, \eg the 2-nd layer respectively merges $\{\bm{s}_1,\bm{s}_2,\bm{s}_3,\bm{s}_4,\bm{s}_5,\bm{s}_6\}$, $\{\bm{s}_7,\bm{s}_8\}$ into two clusters to carry Self-ATT.

To construct an IC and VQA model based on ACF, besides building 1-D ACF for the language decoder, we also extend it to the 2-D ACF as the vision encoder to merge 2-D image patches into bigger ones.
The distinction between IC and VQA lies in their decoder inputs. In VQA, the decoder input is the question, whereas in IC, it is a sentence derived from the previous state.
Moreover, we design two strategies to reduce the cost of calculating the clustering matrix $\bm{C}$ for the 2-D case, which are: (1) using independence assumption to decompose the 2-D distribution into two 1-D calculation, \ie the horizontal and vertical dimension, (2) down-up sampling strategy. By using vision ACF (2-D ACF) as the encoder and language ACF (1-D ACF) as the decoder, the built image-to-text generation model exploits the same inductive bias to discover hidden structures to learn better region-phrase alignments. For example, as Figure~\ref{fig:fig1}(d) shows, the patches of the object ``snow board'' and the phrase ``a snow board'' are respectively adaptively clustered. To sum up, our contributions are:
\begin{itemize}
	\setlength{\itemsep}{3pt}
	\setlength{\parsep}{3pt}
	\setlength{\parskip}{3pt}
        \vspace{-6pt}
	\item We propose a novel \textbf{Ada-ClustFormer} that can adaptively cluster the neighbor elements for carrying Self-ATT to learn global-local contexts.
	\vspace{-6pt}
	\item We design both 1-D and 2-D ACF for building a homogeneous model to transfer more structural commonalities for better captions/answers in IC/VQA tasks.
	\vspace{-6pt}
	\item We propose two strategies which are independence assumption and down-up sampling to reduce the computation burdens of 2-D ACF.%%xxx independence decomposition=assumption
	\vspace{-6pt}
        \item We carry out exhaustive experiments to validate the effectiveness of ACF.

\end{itemize}

\section{Related Works}
\label{sec:related_work}
\noindent\textbf{Image Captioning (IC).} IC aims to generate descriptions according to the given images.
Typically, an encoder-decoder paradigm is used to convert visual inputs to sequence outputs.
Nowadays, Transformer-based models have shown their might and replace RNN-based decoders in IC~\cite{Li_2019_ICCV,herdade2019image,huang2019attention}. Subsequently, more advanced Transformer-based decoders are proposed, \eg $\mathcal{M}^2$ Transformer~\cite{cornia2020meshed} proposes a meshed-memory Transformer to interact with the low-level and high-level features.

%However, these models still use CNN-based feature extractors. 
More recently, witnessing the boom of Vision Transformers (ViT)~\cite{dosovitskiy2020vit}, researchers use ViT-based visual encoders for captioning. Early attempts such as CPTR\cite{liu2021cptr} replace ROI-based features with ViT-extracted grid-based features, while DLCT~\cite{luo2021dual} combines ROI-based and grid-based representations to complement each other’s strengths. Beyond visual encoders, some works leverage knowledge distillation from Vision-Language BERTs~\cite{lee2020vilbertscore} to enhance caption quality. VinVL~\cite{zhang2021vinvl} integrates object detection into the captioning pipeline, whereas EHAT~\cite{10490218} extends the task to cross-lingual captioning within a unified model. For dense captioning, DCMSTRD~\cite{DCMSTRD} introduces a multi-scale module to directly generate region descriptions without explicit matching. Clus~\cite{fayou2024clustering} improves semantic alignment by clustering image and text features into coherent pairs. In addition, reinforcement learning~\cite{rennie2017self,Zheng_fcs} has been employed to refine captions based on sequence-level rewards. Large-scale pretraining has also become a powerful paradigm~\cite{Yongliang_fcs}, as demonstrated by ClipCap~\cite{mokady2021clipcap}, LEMON~\cite{hu2022scaling}, PL-UIC~\cite{zhu2023prompt}, and BLIP-2~\cite{li2023blip}. Notably, most of these approaches adopt ViT~\cite{dosovitskiy2020vit} or Swin Transformer~\cite{liu2021swin} as their backbone.

\noindent\textbf{Visual Question Answering (VQA).}
VQA is another image understanding task that involves answering specific questions related to given images.
Like image captioning, VQA also requires the ability to analyze and comprehend the visual content of the input image. In addition, it needs to comprehend the context of the question.
Therefore, a multimodal fusion architecture is proposed, wherein a CNN is utilized as the visual encoder, and an LSTM or RNN is employed as the language decoder~\cite{antol2015vqa}.
However, there may be a lack of direct correspondence between images and words in the mapping process due to the distinct architectures of the encoder and decoder.
Hence, recent research incorporate the attention mechanism to learn the correspondence between the image and the question/answers adaptively, \eg MCAN~\cite{yu2019deep}, APN~\cite{yang2021auto}, and GQA~\cite{jiang2020defense}.
MCAN~\cite{yu2019deep} introduces a modular co-attention mechanisms that enable the model to focus on relevant regions of the image and words in the question simultaneously.
GQA~\cite{jiang2020defense} introduces a VQA model using grid-based features instead of ROI-based features.
PW-VQA~\cite{PWVQA} introduces a counterfactual inference to mitigate the spurious associations between language and vision.
CroMIC-QA~\cite{CroMIC-QA} leverages the semantics of both the image and the incomplete question to complete the question and generate the corresponding answer.

Among the previous IC and VQA models, Auto-Parsing Network (APN)~\cite{yang2021auto} has a similar motivation as ours, which also inserts a clustering matrix into the Self-ATT layer. APN only considers whether pairwise neighboring elements should be clustered or not, while we calculate this probability from a more global scope. However, our method (ACF) calculates this matrix differently. Specifically, we consider whether the next element is similar to the previous clustered elements. More importantly, we extend our ACF into the 2-D case, which can adaptively cluster the visual patches into regions, while APN only treats a sequence of ROI features as the visual input and still applies a 1-D clustering matrix to address it.

\noindent\textbf{Global-Local Transformer.}
To overcome the fully connected graph prior in standard Transformers, recent studies have explored global-local architectures to model sparse and structured dependencies in language~\cite{luong2015effective,beltagy2020longformer,liang2023open}.
In natural language processing, Global-local~\cite{luong2015effective} adopts a fixed-size combination of global and local attention for neural machine translation, while Longformer~\cite{beltagy2020longformer} combines global attention with local sliding windows to balance inductive bias and long-range context modeling.
Such mechanisms have also proven effective in the vision domain~\cite{chen2021glit,hassani2023neighborhood,liang2023local}. For instance, GLiT~\cite{chen2021glit} adaptively balances global and local information in image features, NAT~\cite{hassani2023neighborhood} expands the receptive field via scalable sliding window attention without pixel shifts, and Graph-S~\cite{Zizhang_fcs} incorporates a graph transformer with boundary-aware attention to jointly capture region-level and pixel-level features.

However, these models are exclusively developed in a single domain (either NLP or CV),
while our ACF provides a general approach in both the vision and language domains.
Thus, using ACF to build the IC and VQA model encourage learning a unified structure space for transferring more structure commonalities.

\section{Methodology}
Compared with the classic Transformer, Ada-ClustFormer (ACF) inserts an adaptively clustering matrix $\bm{C}$ into each Self-Attention (Self-ATT) layer to adaptively control the scope of Self-ATT. The calculation of $\bm{C}$ is detailed in Section~\ref{Matrix_C} where we first show the 1-D language case and then extend it to the 2-D vision case. By stacking these revised Self-ATT layers, ACF can be built for constructing the vision encoder and language decoder for image captioning and VQA (cf. Section~\ref{sec:encoder-decoder}).

\subsection{Ada-ClustFormer}
\label{Matrix_C}

\noindent\textbf{Adaptive Clustering Matrix $\bm{C}$.}
%xxx
ACF is built based on Transformer, whose most elemental building block is the Multi-Head Attention (\textbf{MHA})~\cite{vaswani2017attention}. From the perspective of structure learning~\cite{battaglia2018relational}, single-head Self-ATT constructs a fully-connected (FC) graph where the nodes are the elements (\eg words) and the pairwise edges are weighted by the pairwise attention weight. Correspondingly, a $h$-head Self-ATT constructs $h$ FC graphs with different edge weights. To sparsify this FC-graph, researchers~\cite{dosovitskiy2020vit,liu2021swin} propose to carry Self-ATT in fixed-size windows, which is achieved by revising attention Head $\mathcal{H}$ in MHA:
\begin{equation} \label{equ:cons_head}
\begin{aligned}
 \bm{C}\textbf{-based Head}:  \quad  &\mathcal{H}'=(\mathcal{H}\otimes \bm{C}),\\
\end{aligned}
\end{equation}
where ``$\otimes$'' denotes the element-wise production; $\bm{C}$ is a $N \times N$ \textbf{binary} clustering matrix that only the elements in the window can attend to each other, \ie if the window size is $w$, $\bm{C}_{i,j}=1$ if $|i-j|\le w$ and $\bm{C}_{i,j}=0$ if $|i-j|> w$. However, language or vision data usually have diverse graininess, \eg a phrase may contain different numbers of words or an object may cover diverse spatial regions, while the fixed-size windows cannot capture the varying graininess. 

To amend this, we revise the binary $\bm{C}$ to a \textbf{probabilistic} one where $\bm{C}_{i,j}$ softly determines whether to cluster the embeddings from $\bm{s}_i$ to $\bm{s}_j$ for carrying Self-ATT. Then if $\bm{C}_{i,j}$ is small, the pairwise attention in $\mathcal{A}$ between $\bm{s}_i$ and $\bm{s}_j$ is weakened in Eq.~\eqref{equ:cons_head}, which means $\bm{s}_i$ and $\bm{s}_j$ are less likely to stay in the same cluster. To adaptively decide the window size according to each specific input for capturing the varying graininess, we use the input itself to calculate $\bm{C}_{i,j}$:
\begin{equation} \label{equ:equ_M1}
    \bm{C}_{i,j}=P(\bm{s}_i,...,\bm{s}_j)=\prod_{k=i}^j P(\bm{s}_k|\bm{s}_i,...,\bm{s}_{k-1}),
\end{equation}
where the joint distribution is decomposed to the productions of conditional distributions $P(\bm{s}_k|\bm{s}_i,...,\bm{s}_{k-1})$, which softly decides whether to merge a new element $\bm{s}_k$ into the sub-sequence $\{\bm{s}_i,...,\bm{s}_{k-1}\}$. In the implementation, $P(\bm{s}_k|\bm{s}_i,...,\bm{s}_{k-1})$ is calculated as:
\begin{equation} \label{equ:equ_M2}
    P(\bm{s}_k|\bm{s}_i,...,\bm{s}_{k-1}) = \text{Sigmoid}(\text{FC}([\bm{s}_k,\bm{s}_{i:k-1}])),
\end{equation}
where $\bm{s}_{i:k-1}$ is the mean pooling of the embeddings from $\bm{s}_i$ to $\bm{s}_{k-1}$.
Intuitively, Eq.~\eqref{equ:equ_M2} exploits the context of the whole sub-sequence $\{\bm{s}_i,...,\bm{s}_{k-1}\}$ to decide whether to merge a new element $\{\bm{s}_{k}\}$ into this sub-sequence. 
To model $P(\bm{s}_k|\bm{s}_i,...,\bm{s}_{k-1})$, we employ the sigmoid function, which transforms the raw scores derived from the input sequence into probabilities in $[0,1]$, enabling soft decisions and ensuring differentiability for optimization.
Note that Eq.~\eqref{equ:equ_M1} and Eq.~\eqref{equ:equ_M2} only make sense when $i<k$. Since clustering the embeddings from $\bm{s}_i$ to $\bm{s}_k$ equals to clustering from $\bm{s}_k$ to $\bm{s}_i$, we set $\bm{C}_{i,k} = \bm{C}_{k,i}$ if $i>k$ and since a single element $\bm{s}_i$ is itself a cluster, we set $\bm{C}_{i,i}=1$. 

\begin{figure}[t]
\centering
\includegraphics[width=0.9\linewidth,clip]{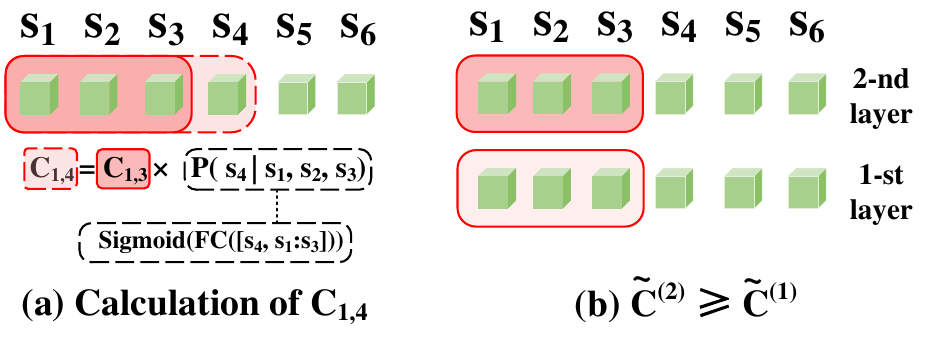}
   \caption{(a) shows how to calculate $C_{1,4}$, where the shade denotes the probability value, the darker the color, the larger the probability value. (b) shows that the clustered elements in the lower layer will be further clustered in a higher layer, \eg the color of $\{\bm{s}_1, \bm{s}_2, \bm{s}_3\}$ in the 2-nd layer is darker than the 1-st layer.
   }
   
\label{fig:fig2}
\vspace{-0.1in}
\end{figure}

%From Eq.~\eqref{equ:equ_M1}, we can also find that:
Let us reconsider Eq.~\eqref{equ:equ_M1}. According to the chain rule of probability, $\bm{C}_{i,j}$ can be rewritten as:
\begin{equation} \label{equ:equ_M3}
\begin{aligned}
 \bm{C}_{i,j}=&P(\bm{s}_j|\bm{s}_i,...,\bm{s}_{j-1})\times P(\bm{s}_i,...,\bm{s}_{j-1}) \\
 =& P(\bm{s}_j|\bm{s}_i,...,\bm{s}_{j-1}) \times \bm{C}_{i,j-1}.
\end{aligned}
\end{equation}
Since $P(\bm{s}_j|\bm{s}_i,...,\bm{s}_{j-1}) \le 1$, we have $\bm{C}_{i,j} \le \bm{C}_{i,j-1}$, which means that two elements in the shorter distance are more likely to be clustered for carrying Self-ATT. In this way, local contexts are encouraged to be captured, as is shown in Figure~\ref{fig:fig2}(a).

\noindent\textbf{Stacking Revised Self-ATT.}
\label{sec:parsing}
To learn global contexts, we can stack these revised Self-ATT layers. When stacking, we hope that the higher layers will carry Self-ATT in bigger windows than the lower layers to capture the global contexts~\cite{wang2019tree,yang2021auto}. To achieve this, for the $m$-th layer, we re-calculate $\bm{C}^{(m)}$ as $\tilde{\bm{C}}^{(m)}$:
\begin{equation}\label{equ:sr}
    \tilde{\bm{C}}^{(m)} = (1- \bm{C}^{(m)})\tilde{\bm{C}}^{(m-1)} + \bm{C}^{(m)}.
\end{equation}
Then $\tilde{\bm{C}}^{(m)}$ is used in Eq.~\eqref{equ:cons_head} when $m>1$ and $\tilde{\bm{C}}^{(1)}={\bm{C}}^{(1)}$.
Since $0 \le \bm{C}^{(m)}_{i,j} \le 1$, $\tilde{\bm{C}}^{(m)}_{i,j}$ is a convex combination of $\tilde{\bm{C}}^{(m-1)}_{i,j}$ and 1, which means that $\tilde{\bm{C}}^{(m-1)}_{i,j} \le \tilde{\bm{C}}^{(m)}_{i,j} \le 1$. If $\tilde{\bm{C}}^{(m-1)}_{i,j}$ is large, \ie the sub-sequence $\{\bm{s}_i,...,\bm{s}_j\}$ should be clustered in the ${(m-1)}$-th layer, then $\tilde{\bm{C}}^{(m)}_{i,j}$ must be larger, \ie $\{\bm{s}_i,...,\bm{s}_j\}$ is also clustered in the ${m}$-th layer. For example, Figure~\ref{fig:fig2}(b) shows that the 2-nd layer will further cluster $\{\bm{s}_1,\bm{s}_2,\bm{s}_3\}$ since $\tilde{\bm{C}}^{(1)}_{1,3} \le \tilde{\bm{C}}^{(2)}_{1,3}$. Thus, the higher layers will carry Self-ATT in a bigger window than the lower layers to learn more global contexts.

\noindent\textbf{2-D Clustering Matrix.}
Eq.~\eqref{equ:equ_M1} shows how to calculate $\bm{C}$ when the input is a 1-D language sequence, next we extend it to the 2-D vision surface. Given a 2-D feature map $\bm{V}=\{\bm{v}_{1,1},...,\bm{v}_{H,W}\}$, we use $\bm{C}_{i,j;x,y}$ to denote the probability that softly decides whether a sub-region $\{\bm{v}_{i,x},...,\bm{v}_{j,y}\}$ should be clustered or not, which is:
\begin{equation} \label{equ:equ_M2D1}
\begin{aligned}
    &\bm{C}_{i,j;x,y} =  P(\bm{v}_{i;x},...,\bm{v}_{j;y})\\
    = & \prod_{k=i}^j \prod_{u=x}^y P(\bm{v}_{k;u}|\bm{v}_{i;x},\bm{v}_{i+1;x},...,\bm{v}_{k-1;u-1})\\
\end{aligned}
\end{equation}
where $i,j$ and $x,y$ respectively denote the horizontal and vertical dimensions. To cover all the sub-regions in a $H\times W$ map, it requires applying $O(H^2\times W^2)$ times for Eq.~\eqref{equ:equ_M2} to get all the probabilities. To reduce the computation burden, we apply the independence assumption to decompose the 2-D distribution into two independent ones, which respectively correspond to the horizontal and vertical dimensions:
\begin{equation} \label{equ:equ_M2D2}
\begin{aligned}
   & P(\bm{v}_{i;x},...,\bm{v}_{j;y})=  P_h(\bm{v}_{i;x},...\bm{v}_{j;x})P_v(\bm{v}_{i;x},...,\bm{v}_{i;y})\\
    = &\prod_{k=i}^j P_h(\bm{v}_{k;x}|\bm{v}_{i;x},...,\bm{v}_{k-1;x})\prod_{u=x}^y P_v(\bm{v}_{i;x}|\bm{v}_{i;x},...,\bm{v}_{i;u-1}),
\end{aligned}
\end{equation}
In this way, we only need to apply $O(H^2+W^2)$ times for Eq.~\eqref{equ:equ_M2} and once matrix production. Noteworthy, as sketched in Figure~\ref{fig:fig2}, for the 2-D region which spans the horizontal axis from $i$ to $j$ and the vertical axis from $x$ to $y$, we use the left-most vertical and top-most horizontal to calculate two 1-D distributions and then multiply them to get $\bm{C}_{i,j;x,y}$.

\iffalse
As Figure~\ref{fig:overview}(a) shows, to calculate $\bm{C}_{1,4;1,3}$, for the vertical distribution $P_v$, the horizontal ordinate is fixed to $1$ and the vertical ordinate changes. $P_h(\bm{v}_{k;1}|\bm{v}_{1;1},...,\bm{v}_{k-1;1})|_{k=1,2,3,4}$ and $P_v(\bm{v}_{1;u}|\bm{v}_{1;1},...,\bm{v}_{1;u-1})|_{u=1,2,3}$ are calculated in the same way as Eq.~\eqref{equ:equ_M2}. The aforementioned symmetric characteristic is also applied.

\begin{figure}[t]
\centering
\includegraphics[width=0.9\linewidth,clip]{figures/overview.pdf}
   \caption{(a) The example of 2-D $\bm{C}$, where $C_{1,4;1,3}$ is used as the example, which is decomposed into vertical and horizontal directions probabilities. (b) Overview of the Down-Up Sampling Strategy.
   }
\label{fig:overview}
\vspace{-0.2in}
\end{figure}

\noindent\textbf{Down-Up Sampling Strategy.}
\label{sec:down-up}
If the sequence (feature map) is too long (big), we can apply the Down-Up Sampling Strategy to reduce the computation cost.
We use a 1-D language case as an example to show this strategy. For $\bm{S}=\{\bm{s}_1,...,\bm{s}_L\}$, we can downsample it to $\bm{\bar{S}}=\{\bm{\bar{s}}_1,...,\bm{\bar{s}}_{L/2}\}$ where $\bm{\bar{s}}_i$ is the mean pooling of $\bm{s}_{2*i-1}$ and $\bm{s}_{2*i}$. Then $\bm{\bar{S}}$ is used in Eq.~\eqref{equ:equ_M1} and Eq.~\eqref{equ:equ_M2} to get $\bm{\bar{C}}$. To upsample $\bar{\bm{C}}$ to the original size, we set $\bm{C}_{i,j}=\bm{\bar{C}}_{\lceil i/2\rceil,\lceil j/2 \rceil }$. Figure~\ref{fig:overview}(b) shows one simple case where $L=4$.
\fi

\noindent\textbf{Expansion on ROI feature.}
\label{sec:ROI}
 The method above applies to the grid-based feature, whose feature map is formed as $H \times W $. For the ROI-based feature, the position of the regions is not certain and the regions are not arranged as grids. Given the n regions: $\{(u_1,v_1),(u_2,v_2),...,(u_n,v_n)\}$, where $u,v$ represents the center coordinates, it is divided into $W$ groups based on $u$. If n cannot be divided by $W$, we fill the last group with dummy regions until it has $H$ regions, so that each group contains $H$ regions.
The dummy regions are filled with a value of 0, ensuring that no additional information is added and that they have no effect on the attention calculation.
Then we sort the regions by $v$ in each group.
Thus, a sorted ROI feature map is obtained with the same form as grid features.

\begin{figure}[t]
\centering
\includegraphics[width=1\linewidth,clip]{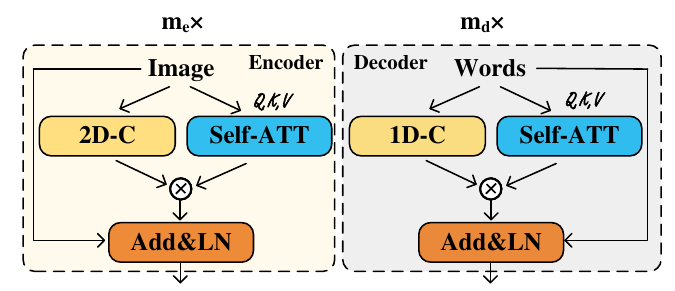}
   \caption{Overview of our ACF-based encoder and decoder.
The ``Add$\&$LN'' is the Add and Layer Normalization.
$m_e$/$m_d$ represent the number of the encoder/decoder layers, respectively. 
   }
   
\label{fig:architecture}
\vspace{-0.1in}
\end{figure}

\iffalse
\begin{figure}[t]
\centering
\includegraphics[width=1\linewidth,clip]{figures/fig6.pdf}
   \caption{Overview of our ACF-based encoder-decoder VQA model.
$m_i$/$m_q$/$m_c$ represent the number of the image encoder/question encoder/cross-modal decoder layers, respectively. 
   }
   
\label{fig:fig6}
\vspace{-0.2in}
\end{figure}
\fi

\subsection{Encoder-Decoder Architecture}
%\noindent\textbf{Clusterformer}
\label{sec:encoder-decoder}
As is shown in Figure~\ref{fig:architecture}, we apply the ACF to build the vision encoder and language decoder for IC and VQA.
Compared to the classic Transformer, our ACF introduces clustering-restrained attention head.
Specifically, in the encoder, we calculate a 2-D clustering matrix $\bm{C}$ (cf. Eq.~\eqref{equ:equ_M2D1}) to softly cluster the elements for carrying Self-ATT. Similarly, in the decoder, the attention head is revised with the 1-D $\bm{C}$ (cf. Eq.~\eqref{equ:equ_M3}).

To further improve the quality of IC, we employ Reinforcement learning (RL)~\cite{rennie2017self}. Specifically, we optimize the model by minimizing the cross-entropy loss and maximizing the RL reward.
First, we train the model by minimizing the cross-entropy loss:
\begin{equation}
     L_{CE} =- \log P(\bm{Z}^*),
 \label{equ:equ_celoss}
\end{equation}
where $\bm{Z}^*$ is the ground-truth captions.
Then, we further train the model by minimizing the negative reward:
\begin{equation}
     L_{rl} = - \mathbb{E}_{\bm{Z}^s \sim P(\bm{Z})}(\mathbb{S}(\bm{Z}^*, \bm{Z}^s)),
 \label{equ:equ_rlloss}
\end{equation}
where $\bm{Z}^s$ is sampled from $\bm{Z}$, $\mathbb{E}$ represents the mathematical expectation, and $\mathbb{S}$ represents the evaluation metrics, \eg CIDEr.

\begin{figure*}[htbp]
\centering
\includegraphics[width=1\linewidth,clip]{figures/fig53.pdf}
   \caption{Examples of the generated captions by BASE and ACF model. We visualize the 2-D $C$ and 1-D $C$ in the 1-st, 3-rd, 5-th, and 6-th layers as the clustered patches. And the ``BASE\_total'' and ``ACF\_total'' represent the normalized fine-grained alignment score evaluated on the whole dataset by the BASE and ACF model, respectively. "GT" represents for the ground-truth label. VQA-v2 dataset provides 10 possible answers for each question. One of the most confident answers of each question is listed in this figure.
   %In detail, $\{ \bm{v}_{i,x},...,\bm{v}_{j,y}\}$ are viewed as a cluster when $\bm{C}_{i,j;x,y} > 0.8$, and $\{\bm{s}_{i},...,\bm{s}_{j}\}$ are viewed as a cluster when $\bm{C}_{i,j} > 0.5$. 
   }
\label{fig:fig5}
\vspace{-0.2in}
\end{figure*}

\section{Image Captioning}

\label{sec:Experiments}
\subsection{Dataset, Metrics, and Settings}
\label{sec:settings}
\noindent\textbf{MSCOCO Dataset.} Following~\cite{pan2020x,yang2021auto,huang2019attention,herdade2019image,cornia2020meshed},
we train and evaluate our model on MSCOCO~\cite{lin2014microsoft}, which contains $123,287$ images, and each one is annotated with 5 captions. In the experiments, we use the Karpathy split (113,287/5,000/5,000 train/val/test images)~\cite{karpathy2015deep} for offline training and the official split (40775 test images) for online testing.

\noindent\textbf{Metrics.}  We adopt five widely-used metrics in captioning for evaluation, including BLEU~\cite{papineni2002bleu}, METEOR~\cite{banerjee2005meteor}, ROUGE-L~\cite{lin2004rouge}, CIDEr~\cite{vedantam2015cider}, and SPICE~\cite{anderson2016spice}.
Besides, we calculate the fine-grained alignment score~\cite{mao2022rethinking, zha2019context,yao2022filip} to evaluate the correspondence of the visual and language patches.

Given the visual feature $D_v$ and the text feature $D_t$, we firstly calculate $V_{score}=D_v \cdot D_t ^T$, and $T_{score}=D_t \cdot D_v ^T$, where ``$\cdot$'' represents the matrix multiplication. Then we count the number of coincidence of the maximum index of $V_{score}$ and $T_{score}$. Finally, we normalize this number and obtain the normalized fine-grained alignment score.

\begin{table}[t]
\begin{center}
\caption{Performance of the ablation models.}
\label{table:ab}
%ROUGE-L
\scalebox{1}{
\begin{tabular}{l cc c c c c c}
		\hline
		   Models  & B@4 & M & R &   C & S\\ \hline
		   \textbf{BASE}&$40.0$ & $29.7$ & $59.6$ & $134.4$ & $23.4$\\
           \textbf{ACF$_{\bm{DE}}$}&$ 40.2$ & $ 29.8$ & $59.9$ & $135.1$ & $23.7$ \\
           \textbf{ACF$_{\bm{EN-2D}}$}&$40.4$ & $29.8$ & $60.0$ & $135.9$ & $23.7$ \\
          \textbf{ACF$_{\bm{EN-1D}}$}&$ 40.0$ & $ 29.4$ & $59.4$ & $134.5$ & $23.2$ \\
           \textbf{w/o-Eq.~\eqref{equ:sr}}&$ 39.1$ & $ 28.7$ & $58.8$ & $132.6$ & $22.8$ \\
           \textbf{w/o DS}&$ 41.2$ & $ 30.0$ & $60.5$  & $137.8$ & $24.2$ \\
           \textbf{DS@4}&$ 39.8$ & $ 29.0$ & $59.3$  & $135.5$ & $23.0$ \\
           \textbf{FS@2}&$ 39.8$ & $ 29.2$ & $59.1$  & $134.9$ & $22.9$\\\hline  
           \textbf{ACF}& $\bm{41.1}$ & $\bm{30.1}$ & $\bm{60.2}$ & $\bm{137.8}$ & $\bm{24.1}$\\  
           \hline

\end{tabular}
}
\end{center}
\vspace{-0.1in}
\end{table}

\begin{table*}[htbp]
\begin{center}
\caption{The performance of SOTA methods on MSCOCO Karpathy split.}
\label{table:tab_sota} 
\scalebox{1}{
\begin{tabular}{lccccccccccc}
\hline
\multirow{2}{*}{Models} & \multicolumn{5}{c}{Cross-Entropy Loss} & &\multicolumn{5}{c}{CIDEr optimization} \\
\cmidrule(r){2-6}  \cmidrule(r){8-12}
                        & \multicolumn{1}{c}{B@4} & \multicolumn{1}{c}{M} & \multicolumn{1}{c}{R} & \multicolumn{1}{c}{C} & \multicolumn{1}{c}{S} & & \multicolumn{1}{c}{B@4} & \multicolumn{1}{c}{M} & \multicolumn{1}{c}{R} & \multicolumn{1}{c}{C} & \multicolumn{1}{c}{S} \\
\hline

            {$^{\color{red!80}{{\text{\textbf{ROI-based feature}}}}}$} & & & & & & & & & \\[-6pt]
           Up-Down~\cite{anderson2018bottom}  &$36.2 $&$27.0$&$56.4$&$113.5$ &$20.3$&& $36.3$ & $27.7$ & $56.9$ & $120.1$ & $21.4$ \\
           ORT~\cite{herdade2019image} &$35.5$&$28.0$&$56.6$&$115.4$&$21.2$&  & $38.6$ & $28.7$ & $58.4$ & $128.3$ & $22.6$ \\
           AoANet~\cite{huang2019attention}  &$37.2$&$28.4$&$57.5$&$119.8$&$21.4$&   & $38.9$ & $29.2$ & $58.8$ & $129.8$ & $22.4$ \\
           $\mathcal{M}^2$ Transformer~\cite{cornia2020meshed}   &-&-&-&-&-&   & $39.1$ & $29.2$ & $58.6$ & $131.2$ & $22.6$ \\ 
            CATT~\cite{yang2021causal} &$37.3$&$28.5$&$57.4$&$119.0$&$21.5$& &$39.4$&$29.3$& $58.9$&$131.7$&$22.8$\\ 
           APN~\cite{yang2021auto} &-&-&-&-&-& & $39.6$ & $29.2$ & $59.1$ & $131.8$ & $23.0$\\
           X-Transformer~\cite{pan2020x}&$\bm{38.2}$&$28.8$&$\bm{58.0}$&$122.0$&$21.9$& &$39.7$&$29.5$& $59.2$&$132.8$&$23.2$\\
           Oscar-B~\cite{li2020oscar} &$36.5$&$\bm{30.3}$&-&$\bm{123.7}$&$\bm{23.9}$& &$40.5$&$29.7$& -&$137.6$&$22.8$\\\hline

           {$^{\color{red!80}{\text{\textbf{Grid-based feature}}}}$} & & & & & & & & & \\[-6pt]
           CPTR~\cite{liu2021cptr}& -&- & - & - & - &&$40.0$&$29.1$&$59.4$&$129.4$ &-\\
        %   APN$^\flat $~\cite{yang2021auto} &-&-&-&-&-& & $39.2$ & $28.8$ & $58.0$ & $127.5$ & $22.4$\\
           APN$^\sharp $~\cite{yang2021auto} &-&-&-&-&-& & $40.1$ & $29.4$ & $59.4$ & $133.2$ & $23.3$\\
           Dual-Global~\cite{xian2022dual} &-&-&-&-&-& &$40.3$&$29.2$&$59.4$&$132.4$&$23.3$\\
           DLCT~\cite{luo2021dual}&-&-&-&-&-&   & $40.8$&$29.9$&$59.8$&$137.5$&$23.3$\\
           PureT-base~\cite{Wang_Xu_Sun_2022}&-&-&-&-&-& &$40.3$&$29.9$&$59.9$&$137.5$&$23.8$\\\hline

           {$^{\color{red!80}{{\text{\textbf{Visual-language BERT pretraining}}}}}$} & & & & & & & & & \\[-6pt]
           \rowcolor{gray!20}
           RSTNet~\cite{zhou2020unified}& -&- & - & - & -&&$40.1$&$28.9$&$59.5$&$135.6$&$23.3$\\
           \rowcolor{gray!20}
           ViTCAP-small~\cite{fang2022injecting}&$35.7$& $28.8$& $57.6$&$121.8$&$ 22.1$&&$40.1$&$29.4$&$59.4$&$133.1$& $23.0$\\
            \rowcolor{gray!20}
            ViTCAP-large~\cite{fang2022injecting}&$36.3$& $29.3$& $ 58.1$&$125.2$&$ 22.6$&&$41.2$&$30.1$&$60.1$&$138.1$& $24.1$\\
             \rowcolor{gray!20}
            VinVL~\cite{zhang2021vinvl}&-& -&  -&-&-&&$40.9$& $30.9$& $ -$&$140.6$&$ 25.1$\\
             \rowcolor{gray!20}
            BLIP-2(opt)~\cite{li2023blip}&-& -&  -&-&-&&$43.7$&-&-&$145.8$&-\\
           \hline
          $\textbf{ACF-ROI} $  & $36.3$ & $28.1$ & $57.7$ & $123.2$ & $21.9$ && $40.3$ &$\bm{29.9}$ & $\bm{60.0}$ & $\bm{138.3}$ & $\bm{23.4}$\\
            $\textbf{ACF-Grid} $  & $\bm{38.1}$ & $\bm{28.8}$ & $\bm{58.4}$ & $\bm{123.8}$ & $\bm{21.8}$ && $\bm{41.1}$ & $\bm{30.1}$ & $\bm{60.2}$ & $\bm{137.8}$ & $\bm{24.1}$\\
          
           \hline
\end{tabular}
}
\end{center}

\end{table*}

\noindent\textbf{Settings.} All the captions are converted to lowercase and exclude words occurring less than 6 times, resulting in a vocabulary of 9,487 words.
Besides training on the grid features, we also try to expand on the ROI features. In detail, we arrange the ROI as a matrix according to the position of ROI.
We adopt Swin Transformer~\cite{liu2021swin} as the visual encoder to extract the grid features, and Oscar~\cite{li2020oscar} to extract the ROI features. The size of the grid feature map is $H \times W = 12 \times 12$, and we apply the Down-Up Sampling 
Strategy with sampling rate 2. In detail, we downsample the features before calculation of the ACF matrix and then upsample to recover it.
For our input feature map $\bm{V}= \{\bm{v}_{1,1},...,\bm{v}_{12,12}\}$, it is downsampled to $\bm{\bar{V}}=\{\bm{\bar{v}}_{1,1},...,\bm{\bar{v}}_{6,6}\}$.
Then we obtain the 2-D clustering matrix $\bm{\bar{C}}= \{ \bm{\bar{C}}_{1,1},..., \bm{\bar{C}}_{6,6} \} $.
Finally, the upsampling is applied that $\bm{C}_{i,j}=\bm{\bar{C}}_{\lceil i/2\rceil,\lceil j/2 \rceil }$ ,when $i,j \in [1,12]$.
For the ROI feature, we set $H = 6$ without the sampling strategy. 
%We train 20/30 epochs in the cross-entropy/RL stage, both with the batch size of 40.
We conduct 20 epochs of training for the cross-entropy stage, followed by 30 epochs of training for the RL stage, both with a batch size of 40.
In the cross-entropy stage,
the learning rate is $5  \times 10^{-5}$/$1 \times  10^{-4}$ and decays by 0.8 per 5 epochs for grid/ROI features.
In the RL stage,
the learning rate is initialized to $5  \times 10^{-6}$/$2 \times 10 ^{-5}$ and we implement the same decay policy for 10 epochs for grid/ROI features.
Then the “Reduce-On-Plateau” strategy is applied with a decay rate of 0.5 and patience of 3.

\subsection{Ablation Studies}
\label{sec:ablation}
We conduct extensive ablations for validating the effectiveness of Ada-ClustFormer (ACF) as follows: 
\textbf{BASE}: we set both the encoder and decoder as the classic Transformer. \textbf{ACF$_{\bm{DE}}$}: the decoder is set to the 1-D ACF (cf. Eq.~\eqref{equ:equ_M3}). \textbf{ACF$_{\bm{EN}}$-2D}: the encoder is set to 2-D ACF (cf. Eq.~\eqref{equ:equ_M2D2}). \textbf{ACF$_{\bm{EN}}$-1D}: the encoder is set to 1-D ACF where the vision tokens are treated as one sequence where the image patches are arranged from top-left to the bottom-right. \textbf{w/o Eq.~\eqref{equ:sr}}: replacing the Stacking Revised Self-ATT with Self-ATT, which means that the relevance between the lower layer and higher layer C no longer exists.
\textbf{w/o DS}: the Down-Up Sampling strategy is removed.
\textbf{DS@4}: we adjust the Down-Up Sampling rate to 4.
\textbf{FS@2}: we use the fixed-size window in ACF where the window size is set to 2.

Table~\ref{table:ab} shows the performance of the ablation models. Firstly, we observe that our ACF achieves the highest score, which proves its effectiveness. Next, we evaluate the effect of each module respectively. 
We compare ACF with ACF$_{\bm{DE}}$, ACF$_{\bm{EN}}$-2D, ACF$_{\bm{EN}}$-1D, it shows that ACF achieves better results than the classic Self-ATT. 
And there is a significant improvement when the encoder and decoder are both ACF, which indicates that the unified structure can transfer more structural commonalities.
Besides, the results in ACF$_{\bm{EN}}$-1D that treating the 2-D vision tokens as a 1-D sequence cannot achieve a good result.
This result proves the necessity of 2-D ACF calculation instead of treating vision and language as equal modalities.
By comparing with ACF and Eq.~\eqref{equ:sr}, it indicates that the convex constraint is necessary in ACF.
It is also in line with the intuition that the higher layers carry more global semantics.
The results of DS@4 imply that the Down-Up sampling strategy is a trade-off of performance and computation burdens and a smaller sampling size improves the performance.
However, removing the Down-Up sampling strategy results in minimal improvements, indicating that setting the sampling rate to 2 is optimal when considering both efficiency and performance.
Comparing FS@2 with ACF, we observe that ACF achieves better performance which validates the effectiveness of adaptively choosing the attention window.
From another perspective, the fixed-size window is a special case of ACF, where the cluster matrix of the adjacent pair is set to 1.\\

\noindent\textbf{Qualitative Results}.
We visualize the hierarchical structures of the image and the generated captions in Figure~\ref{fig:fig5} according to the 2-D and 1-D clustering matrix calculated from the 1-st, 3-rd, 5-th, and 6-th layers in the encoder and decoder. By inspecting the images and captions, we can find that the patches and the words are respectively clustered, \eg in the left part of (a), the words ``sitting on motorcycles'' are clustered into a phrase, and in the right part, the patches in the ``motorcycles'' region are clustered. 
Due to the stacking revised Self-ATT introduced in Eq.~\eqref{equ:sr}, higher layers encompass a larger window, making it more likely for regions and phrases to be clustered.
For ROI features, the words ``a statue of a horse'' are clustered in the right part of (c), and the two regions of the horse statue are clustered.
More importantly, when uniting the image and caption, we can find that structural commonalities are transferred.
Furthermore, the normalized fine-grained alignment scores are listed in Figure~\ref{fig:fig5} to evaluate the image-text alignment.
We observe that ACF can improve the alignment performance in both grid-based features and ROI-based features.
It implies that more structural commonalities can be transferred, benefiting from the unified clustering architecture in ACF.

\begin{table}[ht]
\begin{center}
\caption{The scores on the MSCOCO online test server.}
\label{table:tab_cap_on}
\scalebox{0.7}{
\begin{tabular}{l c c c c c c c c}
		\hline
		\multirow{2}*{Models}   & \multicolumn{2}{c}{B@4} & \multicolumn{2}{c}{M} & \multicolumn{2}{c}{R} &  \multicolumn{2}{c}{C} \\ \cmidrule(r){2-3} \cmidrule(r){4-5} \cmidrule(r){6-7} \cmidrule(r){8-9} 
		   & c5 & c40 & c5 & c40 & c5 & c40 & c5 & c40\\
		    \hline
            %{$^{\color{red!80}{\text{\textbf{Single-model}}}}$} & & & & & & & & \\[-6pt]
           Up-Down~\cite{anderson2018bottom}  & $36.9$  & $68.5$  & $27.6$  & $36.7$  & $57.1$  & $72.4$  & $117.9$  & $120.5$ \\ 
           ETA~\cite{Li_2019_ICCV} & $38.9$ & $70.2$ & $28.6$ & $38.0$ & $58.6$  & $73.9$  & $122.1$  & $124.4$ \\ 
           APN~\cite{yang2021auto}     & $38.9$ & $70.2$ & $28.8$ & $38.0$ & $58.7$ & $73.7$ & $126.3$ & $127.6$\\
           NG-SAN~\cite{guo2020normalized} &  38.8 & 70.2 & 29.0 & 38.4 & 58.7 & 74.0 & 126.3 & 128.6 \\
           Dual-Global~\cite{xian2022dual}&$39.1$& $71.2$& $28.9$&$38.4$&$58.9$& $74.4$ &$126.3$&$129.2$ \\
           AoANet~\cite{huang2019attention} &$39.4$ &$71.2$&$29.1$&$38.5$&$58.9$&$74.5$&$126.9$&$129.6$\\
          
           $\mathcal{M}^2$ Transformer~\cite{cornia2020meshed} &$39.7$ &$72.8$&$29.4$&$39.0$&$59.2$&$74.8$&$129.3$&$132.1$\\
         
            \rowcolor{gray!20}
            RSTNet~\cite{zhou2020unified}&$39.7$&$72.5$&$29.3$&$38.7$&$59.2$&$74.2$&$130.1$&$132.4$\\
           
            %VinVL~\cite{zhang2021vinvl}&$40.4$&$74.9$&$30.6$&$40.8$&$60.4$&$76.8$&$134.7$&$138.7$\\
        
          \hline
           \textbf{ACF-ROI} & $\bm{39.3}$& $\bm{70.7}$ & $\bm{29.4}$ & $\bm{38.7}$ & $\bm{59.1}$ & $\bm{74.2}$ & $\bm{131.0}$ & $\bm{133.1}$ \\
           \textbf{ACF-Grid} & $\bm{39.0}$& $\bm{71.3}$ & $\bm{29.2}$ & $\bm{39.2}$ & $\bm{59.2}$ & $\bm{74.2}$ & $\bm{130.2}$ & $\bm{132.3}$ \\ 
          
           \hline
\end{tabular}
}
\end{center}
\vspace{-0.3in}
\end{table}

\subsection{Comparisons with SOTA}
\label{sec:sota}
\noindent\textbf{Comparing Methods.}
The majority of SOTA models can be categorized into three groups, as shown in Table~\ref{table:tab_sota}, where the top/middle/bottom parts are the ROI-based/grid-based/BERT-based models. Note that for APN, besides reporting the results in their paper~\cite{yang2021auto}, which is got by using ROI-based features, we also report the performance using the same grid-based features as ours, denoted as  ``APN$^\sharp$''. \\

\noindent\textbf{Results.} 
From Table~\ref{table:tab_sota}, we observe that ACF is comparable to most state-of-the-art performance when compared with ROI-based and grid-based models. Moreover, ACF-Grid and ACF-ROI achieve comparable performance with ViTCAP-large~\cite{fang2022injecting} that distills knowledge from multiple datasets. However, we only use the captions from MSCOCO for training. Moreover, compared with APN$^\sharp $~\cite{yang2021auto} which inserts an additional clustering matrix into the Self-ATT layers in the decoder, ACF achieves higher performance since it inserts the clustering matrix in both vision encoder and language decoder to build a homogeneous model.

Also, we submit the single-model results to the online server~\cite{codalab_competitions} for testing, which is shown in Table~\ref{table:tab_cap_on}. We can see that
ACF achieves the best performance than the other models, even we do not ensemble the results as AoANet~\cite{huang2019attention} and $\mathcal{M}^2$ Transformer~\cite{cornia2020meshed}. And we outperform the large-scale model RSTNet~\cite{zhou2020unified} in most of the metrics.

\section{VQA}

\label{sec:VQA}
\subsection{Dataset, Metrics, and Settings}
\label{sec:vqa_settings}
\noindent\textbf{VQA-v2 Dataset.} Following ~\cite{yang2021auto,yu2019deep,nguyen2018improved}, our VQA model is trained and evaluated on the VQA-v2 dataset. It adheres to the official MSCOCO split, with 82,783/40,504 images for training/offline validation.
Moreover, it includes test-std and test-dev splits for online testing, each with 81,434 images.

\noindent\textbf{Metrics.} The questions in VQA-v2 are categorized into three types: ``yes/no'', ``number'', and ``other''. The accuracy scores for each question type are calculated separately, and then an overall accuracy score is determined. A higher score indicates better performance of the VQA model.

\noindent\textbf{Settings.} 
The encoder settings remain same as the image captioning (cf. Section~\ref{sec:settings}). Follow previous researches~\cite{yang2021auto,yu2019deep}, the question embedding and the multi-modal embedding of the decoder are set to $512$ and $1,024$, respectively. The number of layers is set to 6. The learning rate is initialized to $2.5 \times e^{-5}$ and increases by $2.5 \times e^{-5}$ for the first 4 epochs. Afterward, the learning rate decays by a factor of 0.2 every 2 epochs and the training continues for 10 epochs. The training is performed with a batch size of 64. Similar to IC, we also evaluate the performance of the grid-based features and the ROI-based features on VQA.
%Due to the fact that VQA training does not require iteration as IC, the model training consumes less GPU memory. Therefore, we do not apply the Down-Up sampling strategy to it.
%We still evaluate the Down-Up sampling strategy in our ablation experiments.

\subsection{Ablations}
We also conduct ablation experiments in the context of VQA, following the ablation settings used in IC (cf. Section~\ref{sec:ablation}). 
The results of the offline evaluation can be found in Table~\ref{table:tab_vqa_ab}. By comparing ACF$_{DE}$ and ACF$_{EN}$ with ACF, we observe that the combination of ACF in both the encoder and decoder leads to better question understanding and image comprehension. As a result, the model can provide more accurate answers. The performance of "DS@4" is significantly worse than ACF, even worse than the baseline model. This outcome indicates that reducing the resolution of the image has a substantial impact on VQA performance. Thus, as mentioned earlier, the ACF model does not utilize the Down-Up sampling strategy. The results of "w/o-Eq.~\eqref{equ:sr}" indicates that the stacking revised strategy is also beneficial for VQA.

\noindent\textbf{Qualitative Results}. We present some examples of the VQA results in Figure~\ref{fig:fig5}. We observe that when dealing with the "Number" type question using the grid-based feature, such as Q2 in Figure~\ref{fig:fig5}(a), becomes challenging. It might be attributed to the inherent characteristics of the grid-based feature, as corroborated by GQA~\cite{jiang2020defense}. However, when confronted with other types of questions, such as Q2 in Figure~\ref{fig:fig5}(b), ACF outperforms by providing more detailed answers compared to the baseline.

\begin{table}[t]
\begin{center}
\caption{VQA-v2 Val accuracy scores of ablative baselines.}
\label{table:tab_vqa_ab}
\scalebox{1.00}{
\begin{tabular}{l c c c c}
		\hline
		    Models & Yes/No & Number & Other & Overall  \\ \hline
		    BASE   & $77.45$  & $44.15$ & $58.22$ & $63.58$\\  
		     \textbf{ACF$_{\bm{DE}}$}  & $77.97$  & $44.19$ & $58.61$ & $63.99$\\ 
                \textbf{ACF$_{\bm{EN}}$}&$83.55$  & $44.01$ & $58.93$ & $66.21$\\ 
                 \textbf{DS@4}&$68.22$  & $30.42$ & $39.25$ & $48.99$\\ 
                  \textbf{w/o-Eq.~\eqref{equ:sr}}&$82.22$  & $43.89$ & $58.89$ & $65.68$\\ 
		   \textbf{ACF}  & $\mathbf{84.99}$  & $\mathbf{44.40}$ & $\mathbf{59.26}$ & $\mathbf{66.98}$\\ 
		   \hline
\end{tabular}
}
\end{center}

\end{table}

\begin{table}[t]
\begin{center}
\caption{VQA-v2 test-dev and test-std accuracy of various models.}
\label{table:tab_vqa_st}
\scalebox{0.9}{
\begin{tabular}{l c c c c c}
		\hline

		    \multirow{2}*{Models}  & \multicolumn{4}{c}{Test-dev}  & Test-std\\ \cmidrule(r){2-5} \cmidrule(r){6-6} 
		   & Yes/No & Number & Other & Overall  & Overall \\
		    \hline
		  DCN~\cite{nguyen2018improved}  & $83.51$  & $46.61$ & $57.26$ &$66.87$& $66.97$ \\ 
		  BAN~\cite{kim2018bilinear}  & $85.42$  & $\bm{54.04}$ & $60.52$ &$70.04$& $70.35$ \\ 
		   MCAN~\cite{yu2019deep}  & $85.82$  & $53.26$ & $60.72$ &$70.63$ &$70.90$\\ 
		      PW-VQA~\cite{PWVQA}&$81.80$& $43.90$& $53.01$& $62.63$&-\\
		   APN~\cite{yang2021auto}& $\bm{87.44}$  & $52.68$ & $61.18$ &$71.14$ &$71.33$\\  
     APN$^\sharp$~\cite{yang2021auto}& $87.44$  & $48.05$ & $62.48$ &$71.11$ &$71.21$\\\hline
     \rowcolor{gray!20}
    GQA~\cite{jiang2020defense}&$\bm{89.18}$  & $\bm{58.01}$ & $\bm{64.77}$ &${74.16}$ &$72.71$\\

    \rowcolor{gray!20}
       BLIP-2(opt)~\cite{li2023blip}&-  & -& - &- &${62.3}$\\ 
       \rowcolor{gray!20}
       OSCAR~\cite{li2020oscar}&-  & -& - &- &$\bm{73.82}$\\		  
        \rowcolor{gray!20}
       VK-OOD~\cite{wang2023implicit}&-  & -& - &$\bm{74.8}$&- \\ 
        \hline
                ACF-Grid & $87.31$  & $48.90$ & $\bm{62.50}$ &$\bm{71.18}$ &$\bm{71.37}$\\
                ACF-ROI & $\bm{88.92}$  & $\bm{55.01}$ & $\bm{62.22}$ &$\bm{72.60}$ &$\bm{72.03}$\\
            \hline 
\end{tabular}
}
\end{center}
\vspace{-0.2in}
\end{table}

\subsection{Comparisons with SOTA}
We compare our ACF model on VQA with the Transformer-based state-of-the-art models. The results can be found in Table~\ref{table:tab_vqa_st}.
It shows that our model achieves higher accuracy among the models without large-scale pretraining. When compared with the methods that utilize ROI-based features, the ACF-Grid demonstrates improved performance in the "Yes/No" and "Other" question types. However, for the "Number" question type, our ACF-Grid generally performs worse than the method using ROI-based features. Despite this, our method still outperforms the APN$^\sharp$, which also adopt the grid-based features. Our method closely aligns with the results of GQA~\cite{jiang2020defense}, BLIP-2(opt)~\cite{li2023blip}, OSCAR~\cite{li2020oscar}, and VK-OOD~\cite{wang2023implicit} in Table~\ref{table:tab_vqa_st}, given that they are large-scale pre-trained.

Among these methods, Auto-Parsing Network (APN)~\cite{yang2021auto} has a similar motivation as ours.
However, our ACF has a different approach to obtaining the hidden structure.
APN calculates the distribution pairwise.
For example, APN considers whether $\bm{s}_j$ and its left element $\bm{s}_{j-1}$, $\bm{s}_j$ and its right element $\bm{s}_{j+1}$ are in the same cluster.
However, we calculate the distribution $P$ globally.
In detail, we consider whether $\bm{s}_j$ and the sub-sequence $\{ \bm{s}_{i},..., \bm{s}_{j-1} \}$ are in the same cluster.
Intuitively, we obtain a more comprehensive clustering structure when calculating globally.
Besides, to compare the efficiency, we list the computational demand of ACF, including parameter size and time consumption in Table~\ref{tab:time}. 
The time consumption denotes the training time of 100 batches with a batch size of 10 on a single NVIDIA V100 GPU.
The baseline model and APN~\cite{yang2021auto} are also added for comparison.
The results of No.4-6 suggest that the addition of the independence assumption and down-up sampling significantly exceed the training.
From the perspective of the parameters, ACF (No.6) did not introduce a significant number of parameters compared to the baseline (No.1).
In terms of training time, compared to APN (No.2, which is also 1D), ACF-1D (No.3) exhibits advantages in both efficiency and performance.
In addition to efficiency considerations, our ACF has the advantage of being applicable to both 1D and 2D cases, unlike APN.

\begin{table}[htbp]
\centering
\caption{Computation demand of ACF}

\label{tab:time}
\scalebox{1.00}{
\begin{tabular}{|l|l|l|l|} 
\hline
No. &Model & Para.(M)&  Time(s) \\ 
\hline
1 &baseline & 35.8& 56 \\\hline
2 &APN & 39.7& 89 \\\hline
3 &ACF-1D  & 38.9&  79\\\hline
4 &w/o. DS & 42.1& 183 \\\hline
5 &w/o. Eq.~\eqref{equ:sr} &74.9 &  586\\\hline
6 &ACF &40.5&  129\\
\hline
\end{tabular}
}
\end{table}

\section{Conclusion and Future Work}

We propose a novel global-local Transformer named as Ada-ClustFormer (ACF) that can adaptively cluster the input elements for carrying Self-Attention (Self-ATT) to learn global-local contexts. Specifically, this is achieved by inserting a clustering matrix into the Self-ATT layer, where the probability terms are calculated from the input data and thus ACF can adaptively cluster the elements. Moreover, we use ACF to build an image captioning model and a VQA model to transfer more structural commonalities for better captions. The experiment results confirm the effectiveness of ACF. 
In the future, the ACF can be extended to tackle additional multi-modal tasks beyond IC and VQA.

\section*{Acknowledgments}
We thank the Big Data Computing Center of Southeast University.

{
\bibliographystyle{ieeetr}
\bibliography{egbib}
}

\section{Biography}
\vspace{-33pt}

\begin{IEEEbiography}[{\includegraphics[width=1in,height=1.25in,clip,keepaspectratio]{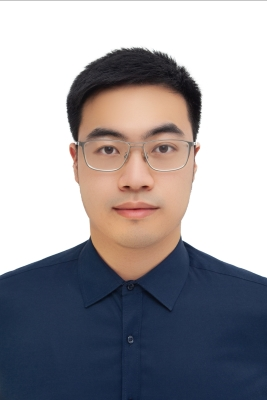}}]{Zihua Wang}
received the BEng degree in telecommunications engineering from University of science and technology of China, China. He is currently a PhD student with Southeast University, China. He is also interning at Tongyi Lab, Alibaba Group. His research interests include computer vision and multi-modal understanding.
\end{IEEEbiography}

\begin{IEEEbiography}[{\includegraphics[width=1in,height=1.25in,clip,keepaspectratio]{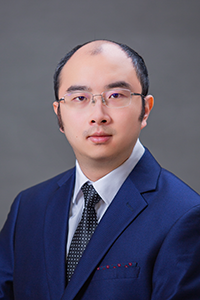}}]{Xu Yang} received the BEng degree in communication engineering from the Nanjing University of Posts and Telecommunications, in 2013, the MEng degree in information processing from Southeast University, in 2016, and the PhD degree in computer science from Nanyang Technological University, in 2021. He is currently an associate professor with the School of Computer Science and Engineering of Southeast University, China. His research interests mainly include computer vision, machine learning, and image captioning.
\end{IEEEbiography}

\begin{IEEEbiography}[{\includegraphics[width=1in,height=1.25in,clip,keepaspectratio]{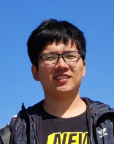}}]{Haiyang Xu} received the MEng degree in in computer science from Southeast University, China. He is currently a Senior Algorithm Expert at Tongyi Lab, Alibaba Group. His research interests include pre-trained language models, multi-modal understanding, and question answering and dialog.
\end{IEEEbiography}
\begin{IEEEbiography}[{\includegraphics[width=1in,height=1.25in,clip,keepaspectratio]{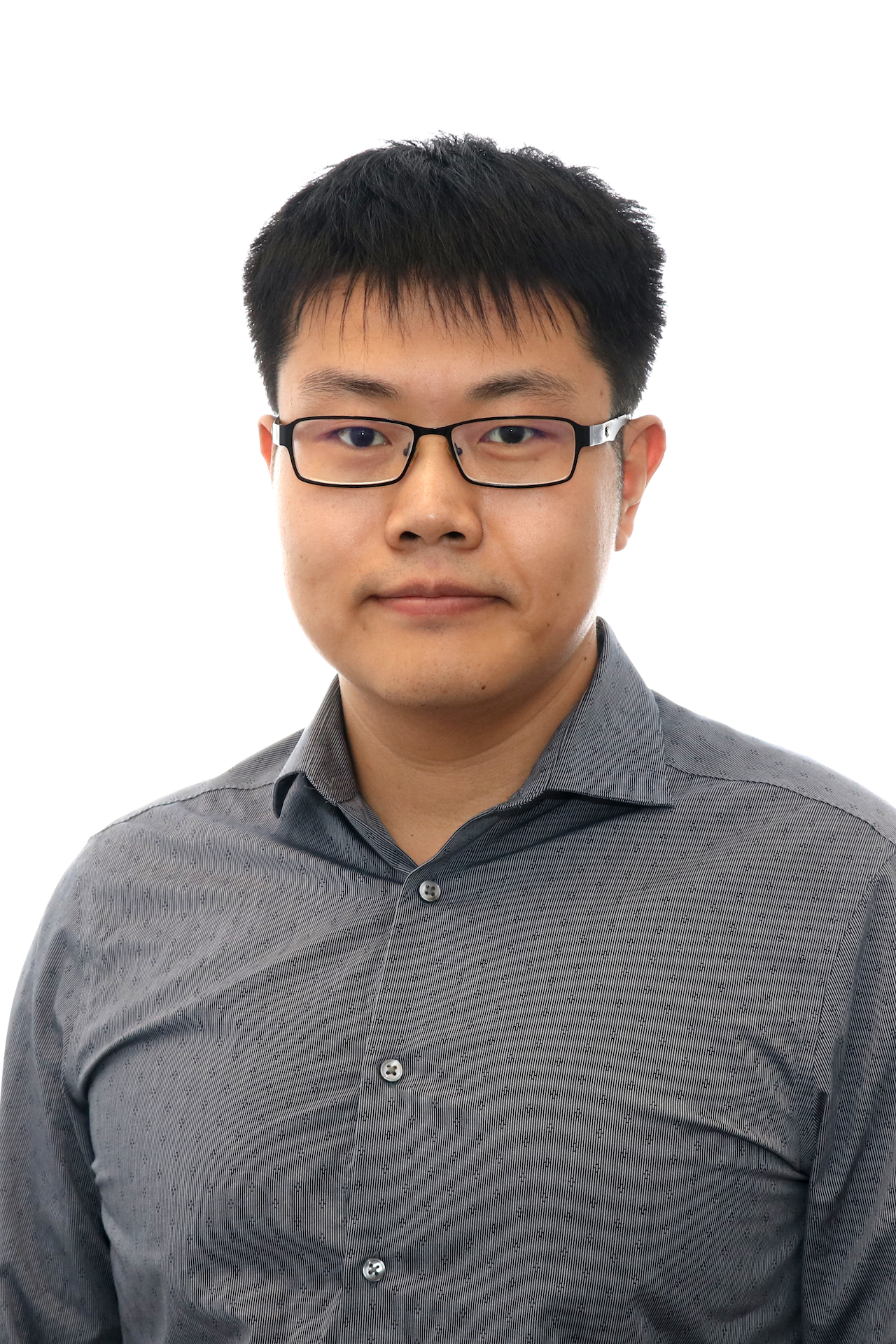}}]{Hanwang Zhang} received the BEng degree in computer science from Zhejiang University, Hangzhou, China in 2009, and the PhD degree in computer science from the National University of Singapore, in 2014. He is currently an assistant professor with the School of Computer Science and Engineering, Nanyang Technological University, Singapore. He was a research scientist with the Department of Computer Science, Columbia University, USA.
\end{IEEEbiography}
\begin{IEEEbiography}[{\includegraphics[width=1in,height=1.25in,clip,keepaspectratio]{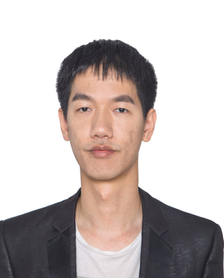}}]{Ming Yan} received the Ph.D. degree from the Institute of Automation, Chinese Academy of Sciences, Beijing, China, in 2016. He is currently a Senior Algorithm Expert at Tongyi Lab, Alibaba Group. His research interests include pre-trained language models, multi-modal understanding, and question answering and dialog.
\end{IEEEbiography}

\begin{IEEEbiography}[{\includegraphics[width=1in,height=1.25in,clip,keepaspectratio]{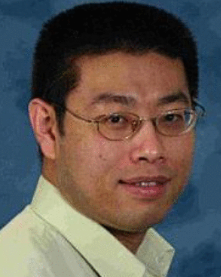}}]{Fei Huang} is currently a Principal Researcher with Tongyi Lab, Alibaba Group. He leads R\&D on NLP foundational technologies, dialogue, and machine translation. His team develops various NLP technologies ranging from lexical, syntactical, semantic, discourse, and deep learning-based algorithms, and integrate them into the Alibaba NLP platform, which supports several hundred internal and external clients with advanced NLP models, systems, and solutions in various industries.
\end{IEEEbiography}

\begin{IEEEbiography}[{\includegraphics[width=1in,height=1.25in,clip,keepaspectratio]{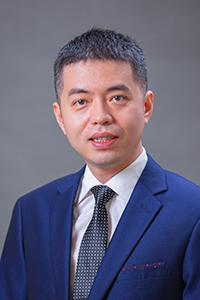}}]{Yu Zhang}
received the B.S. and M.S. degrees in telecommunications engineering from Xidian University, China, and the Ph.D. degree in computer
engineering from Nanyang Technological University, Singapore. He was a Post-Doctoral Fellow with the Bioinformatics Institute, Agency for Science,
Technology and Research, Singapore. He is currently an Associate Professor with Southeast University, China. His research interests include computer
vision.
\end{IEEEbiography}

\vfill

\end{document}